\documentclass[letterpaper,10pt,conference]{ieeeconf}
\IEEEoverridecommandlockouts
\usepackage[T1]{fontenc}
\usepackage{mathptmx}
\usepackage{amsmath,amssymb}
\usepackage{graphicx}
\usepackage{float}
\usepackage{booktabs}
\usepackage{xcolor}
\usepackage{url}
\usepackage[hidelinks]{hyperref}
\usepackage{cite}
\usepackage{tikz}
\usetikzlibrary{arrows.meta,positioning,calc}
\definecolor{geom}{HTML}{E3EDF7}
\definecolor{matl}{HTML}{E6F1E7}
\definecolor{robot}{HTML}{FAE9D7}
\definecolor{ink}{HTML}{233747}
\newcommand{\topic}[1]{\par\smallskip\noindent\textbf{#1}\ }

\DeclareSymbolFont{uprightlatin}{T1}{ptm}{m}{n}
\SetSymbolFont{uprightlatin}{bold}{T1}{ptm}{b}{n}
\DeclareMathSymbol{A}{\mathalpha}{uprightlatin}{`A}
\DeclareMathSymbol{B}{\mathalpha}{uprightlatin}{`B}
\DeclareMathSymbol{C}{\mathalpha}{uprightlatin}{`C}
\DeclareMathSymbol{D}{\mathalpha}{uprightlatin}{`D}
\DeclareMathSymbol{E}{\mathalpha}{uprightlatin}{`E}
\DeclareMathSymbol{F}{\mathalpha}{uprightlatin}{`F}
\DeclareMathSymbol{G}{\mathalpha}{uprightlatin}{`G}
\DeclareMathSymbol{H}{\mathalpha}{uprightlatin}{`H}
\DeclareMathSymbol{I}{\mathalpha}{uprightlatin}{`I}
\DeclareMathSymbol{J}{\mathalpha}{uprightlatin}{`J}
\DeclareMathSymbol{K}{\mathalpha}{uprightlatin}{`K}
\DeclareMathSymbol{L}{\mathalpha}{uprightlatin}{`L}
\DeclareMathSymbol{M}{\mathalpha}{uprightlatin}{`M}
\DeclareMathSymbol{N}{\mathalpha}{uprightlatin}{`N}
\DeclareMathSymbol{O}{\mathalpha}{uprightlatin}{`O}
\DeclareMathSymbol{P}{\mathalpha}{uprightlatin}{`P}
\DeclareMathSymbol{Q}{\mathalpha}{uprightlatin}{`Q}
\DeclareMathSymbol{R}{\mathalpha}{uprightlatin}{`R}
\DeclareMathSymbol{S}{\mathalpha}{uprightlatin}{`S}
\DeclareMathSymbol{T}{\mathalpha}{uprightlatin}{`T}
\DeclareMathSymbol{U}{\mathalpha}{uprightlatin}{`U}
\DeclareMathSymbol{V}{\mathalpha}{uprightlatin}{`V}
\DeclareMathSymbol{W}{\mathalpha}{uprightlatin}{`W}
\DeclareMathSymbol{X}{\mathalpha}{uprightlatin}{`X}
\DeclareMathSymbol{Y}{\mathalpha}{uprightlatin}{`Y}
\DeclareMathSymbol{Z}{\mathalpha}{uprightlatin}{`Z}
\DeclareMathSymbol{a}{\mathalpha}{uprightlatin}{`a}
\DeclareMathSymbol{b}{\mathalpha}{uprightlatin}{`b}
\DeclareMathSymbol{c}{\mathalpha}{uprightlatin}{`c}
\DeclareMathSymbol{d}{\mathalpha}{uprightlatin}{`d}
\DeclareMathSymbol{e}{\mathalpha}{uprightlatin}{`e}
\DeclareMathSymbol{f}{\mathalpha}{uprightlatin}{`f}
\DeclareMathSymbol{g}{\mathalpha}{uprightlatin}{`g}
\DeclareMathSymbol{h}{\mathalpha}{uprightlatin}{`h}
\DeclareMathSymbol{i}{\mathalpha}{uprightlatin}{`i}
\DeclareMathSymbol{j}{\mathalpha}{uprightlatin}{`j}
\DeclareMathSymbol{k}{\mathalpha}{uprightlatin}{`k}
\DeclareMathSymbol{l}{\mathalpha}{uprightlatin}{`l}
\DeclareMathSymbol{m}{\mathalpha}{uprightlatin}{`m}
\DeclareMathSymbol{n}{\mathalpha}{uprightlatin}{`n}
\DeclareMathSymbol{o}{\mathalpha}{uprightlatin}{`o}
\DeclareMathSymbol{p}{\mathalpha}{uprightlatin}{`p}
\DeclareMathSymbol{q}{\mathalpha}{uprightlatin}{`q}
\DeclareMathSymbol{r}{\mathalpha}{uprightlatin}{`r}
\DeclareMathSymbol{s}{\mathalpha}{uprightlatin}{`s}
\DeclareMathSymbol{t}{\mathalpha}{uprightlatin}{`t}
\DeclareMathSymbol{u}{\mathalpha}{uprightlatin}{`u}
\DeclareMathSymbol{v}{\mathalpha}{uprightlatin}{`v}
\DeclareMathSymbol{w}{\mathalpha}{uprightlatin}{`w}
\DeclareMathSymbol{x}{\mathalpha}{uprightlatin}{`x}
\DeclareMathSymbol{y}{\mathalpha}{uprightlatin}{`y}
\DeclareMathSymbol{z}{\mathalpha}{uprightlatin}{`z}

\title{\LARGE\bf DeformSmith: Physics Harness-Guided Hierarchical Generation\\of
Deformable Assets for Robot Manipulation}
\author{Can Li$^{1}$, Jie Gu$^{2}$, Zishun Deng$^{1}$, Jingmin Chen$^{2}$, Lei Sun$^{1}$\\[0.5em]
\normalsize $^{1}$Nankai University \qquad $^{2}$Rightly Robotics, A4x\\[0.4em]
\small\href{https://can-lee.github.io/DeformSmith_web/}{\textcolor{magenta}{\textbf{https://can-lee.github.io/DeformSmith\_web/}}}}

\makeatletter
\IEEEaftertitletext{%
  \begin{minipage}{\textwidth}
    \centering
\includegraphics[width=.96\textwidth]{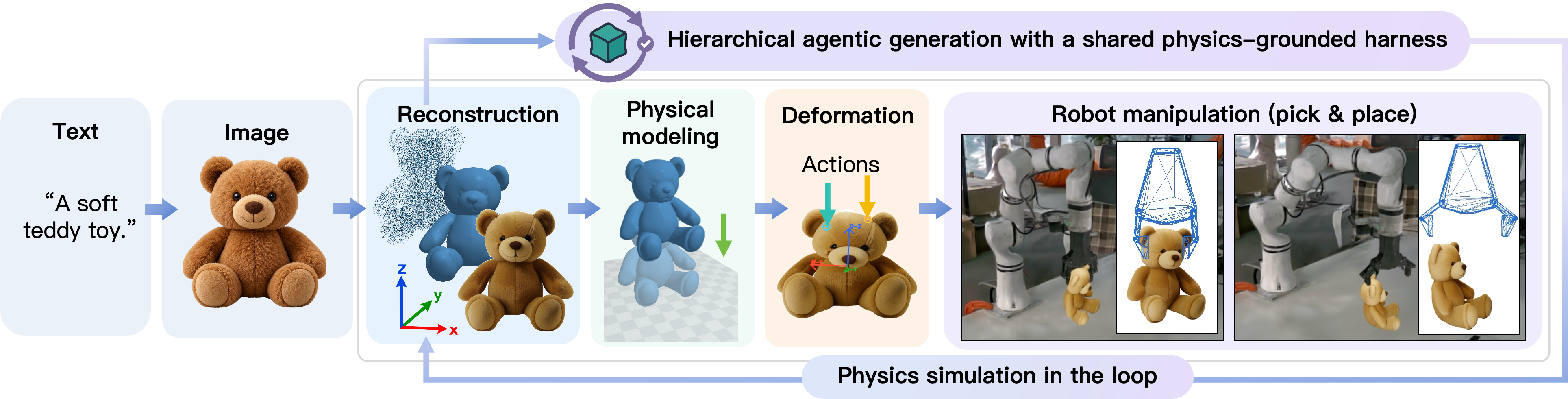}\endgraf
    \nointerlineskip
    \setlength{\abovecaptionskip}{6pt}
    \def\@captype{figure}
    \caption{\textbf{DeformSmith: from text to deformable assets.}
    Generating deformable objects for robot manipulation requires both plausible
    geometry and physical behavior. DeformSmith addresses this
    need through hierarchical agentic generation with a shared physics-grounded
    harness, with physical probes and simulated robot manipulation in the loop
    to guide asset construction and refinement. (Manipulation is rendered using dynamic Gaussian splatting.)}
    \label{fig:teaser}
  \end{minipage}\vspace{1\baselineskip}
}
\makeatother
\begin{document}
\bstctlcite{IEEEcontrol}
\maketitle
\thispagestyle{empty}
\pagestyle{empty}

\begin{abstract}
Creating deformable assets for robot manipulation requires jointly specifying
their geometry, appearance, and physical properties. This is especially challenging for deformable objects, since text and images
provide limited evidence about how they deform and respond to contact, yet
these responses directly affect their suitability for interaction. Automated generation therefore needs
to resolve coupled physical requirements and use interaction evidence to guide
construction and refinement.
We present DeformSmith, a framework that enables automated generation of
interactive, physically credible deformable assets from text or a single image.
Through hierarchical agentic construction and a shared physics-grounded harness,
it progressively builds, tests, and refines geometry, physical models,
material behavior, and robot interaction until the resulting asset is ready
for simulation and manipulation.
Robot interaction closes the generation loop through manipulation feedback and
replayable interaction data. Results show that DeformSmith generates assets
with better visual quality and physical plausibility than state-of-the-art
baselines, including PhysGen3D, PhysGM, and PhysX-Omni, while supporting
the synthesis of data for robotic manipulation of deformable objects.
\end{abstract}

\section{Introduction}
Digital assets provide the objects that populate virtual environments for
visualization, interactive simulation, and embodied applications such as robot
manipulation. Supporting these applications requires diverse assets that
capture both object appearance and physical behavior under interaction.
This requirement is especially challenging for deformable objects, whose
high-dimensional deformation states complicate the joint modeling of geometry,
material properties, and contact interactions. In robot manipulation, for
example, deformation affects whether an object can be grasped, transported,
and released successfully. Generating such assets from text or a single image
could broaden the range of objects available for simulation without measuring
each object in advance. Realizing this goal requires an automated construction
process that turns limited visual and semantic cues into interactive assets,
establishes their physical behavior, and tests their readiness for simulation
and manipulation.

Video-based methods such as PhysTwin~\cite{phystwin}, EMPM~\cite{empm}, and
DeformMaster~\cite{deformmaster} estimate physical parameters from observed
object deformation, while Scalable Real2Sim~\cite{scalable} acquires physical
properties through robot interaction. These approaches rely on observations
of the physical object during motion or interaction. From a single image,
PhysGen3D~\cite{physgen} and PhysGM~\cite{physgm} infer interactive physical
representations without requiring such observations. However, a static image
or a text description does not directly reveal the underlying material parameters
or contact properties needed to simulate an object's response to interaction. In asset construction from text or a single image, inferred
physical properties therefore serve as an initial estimate that must be tested
and refined through simulation. This motivates incorporating physical testing
and interaction feedback into the generation process to guide decisions that
the input alone cannot resolve.

This generation problem involves coupled decisions. Geometry, physical
properties, and contact conditions jointly shape an asset's deformation and
interaction behavior and must therefore be configured and evaluated together.
This motivates a structured generation process that progressively integrates
these components and uses simulation feedback to guide refinement. Agentic generation offers useful foundations:
SceneSmith~\cite{scenesmith} hierarchically organizes scene construction, while the Scientific Generative Agent~\cite{sga} uses simulation to
refine model hypotheses. For deformable assets, these ideas motivate a
hierarchy that establishes each physical prerequisite before subsequent
decisions, together with a shared evaluation and revision process that
preserves consistency as the asset evolves.

A further challenge is to translate physical evaluation into guidance for
generation across the hierarchy. This involves identifying which aspects of an
asset need refinement and assessing whether revisions improve its behavior
during interaction. Because a revision at one stage can affect the behavior
established at others, useful feedback also needs to account for dependencies
across the hierarchy.

To integrate hierarchical generation, physical evaluation, and manipulation
feedback, we present DeformSmith, a framework for automated generation of
interactive, physically credible deformable assets from text or a single image (Fig.~\ref{fig:teaser}). Hierarchical
agentic construction progressively builds, tests, and refines geometry,
physical models, material behavior, and robot interaction toward an
asset ready for simulation and manipulation. A shared physics-grounded harness
connects these stages through a common evaluation and revision process,
using simulation evidence to guide updates while preserving established
physical requirements. Robot manipulation brings the intended use into the
generation loop: manipulation feedback guides further refinement beyond basic
physical probes, and recorded interactions provide replayable manipulation
data. To summarize, our contributions are:
\begin{itemize}
\item We introduce DeformSmith, a fully automated hierarchical agentic
  framework that transforms text or a single image into interactive, physically
  credible deformable assets. DeformSmith progressively constructs and validates the coupled
  geometry, physical models, material behavior, and robot-interaction
  readiness required by deformable objects.
\item We develop a shared physics-grounded harness that couples agentic
  generation with simulation-based evaluation and revision across the hierarchy,
  enabling physically informed refinement while maintaining consistency across
  successive asset updates.
\item We incorporate robot manipulation into the asset-generation loop,
  using manipulation feedback to further validate and refine assets
  beyond basic physical probes while producing replayable manipulation data.
\end{itemize}

\section{Related Work}
\topic{Simulation-ready asset generation.}
Holodeck~\cite{holodeck} builds embodied environments through asset selection and
spatial constraints, while Gen2Sim~\cite{gen2sim} generates simulation assets and
associated tasks. RoboGen~\cite{robogen} automates a propose--generate--learn
cycle. SimFoundry~\cite{simfoundry} reconstructs simulation-ready scenes from
videos and generates object, scene, and task variations for policy learning and
evaluation. SceneSmith and SAGE~\cite{scenesmith,sage} use agentic scene refinement,
while the Scientific Generative Agent~\cite{sga} combines language-model
hypotheses with differentiable simulation for model discovery. These systems automate scene,
task, or model construction; DeformSmith instead coordinates geometry, physics, material,
and robot-interaction decisions for individual deformable assets.

\topic{Physics-informed reconstruction and generation.}
From static inputs, SOPHY~\cite{sophy} generates geometry, appearance, and
physical materials; PhysX-3D~\cite{physx3d} predicts physical properties alongside
geometry; and physically compatible modeling~\cite{physcomp} enforces static
equilibrium. PhysGen3D and PhysGM~\cite{physgen,physgm} infer interactive physical
representations. PhysGaussian~\cite{physgaussian} couples Gaussians with continuum
simulation, while PhysDreamer~\cite{physdreamer} distills video-generation priors
into interactive dynamics. Motion-based methods use observed dynamics:
PAC-NeRF~\cite{pacnerf} estimates continuum parameters; Spring-Gaus and
PhysTwin~\cite{springgaus,phystwin} combine spring-mass dynamics with Gaussian
appearance; and DeformMaster~\cite{deformmaster} learns a physics-neural model
from interaction videos. Scalable Real2Sim~\cite{scalable} uses robotic
pick-and-place to acquire visual and collision geometry and inertial properties.
These methods either infer physical priors from static inputs or rely on observed motion.
DeformSmith revises text- or image-generated assets through simulation probes
and simulated robot pick-and-place.

\begin{figure*}[t]
\centering
\vspace*{2.5mm}
\includegraphics[width=.98\textwidth]{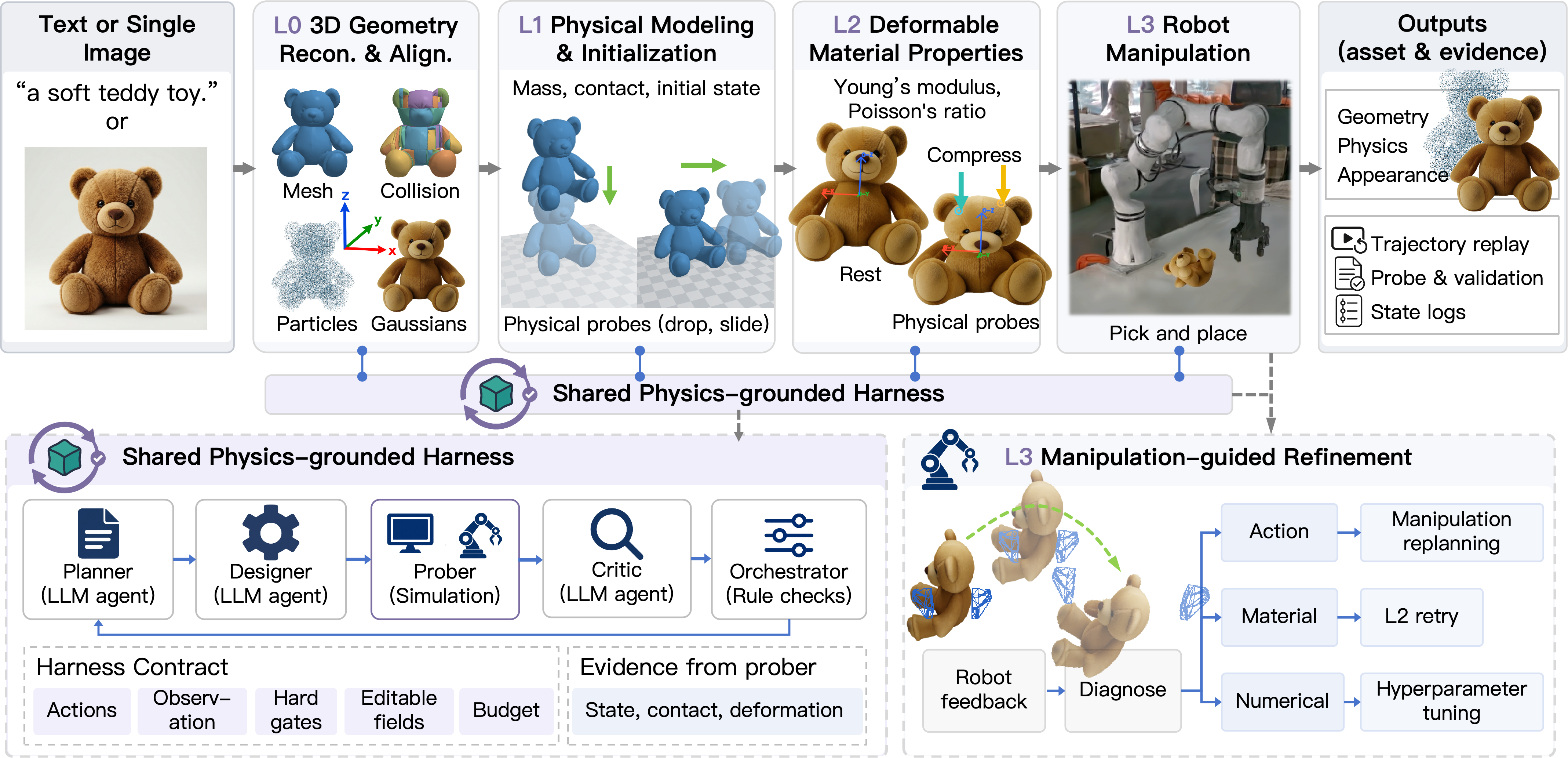}
\caption{\textbf{Overview of DeformSmith.} DeformSmith progressively constructs
deformable assets from text or a single image, with each layer building on the
outputs of the preceding layers. The framework reconstructs 3D geometry,
establishes physical models, configures material properties, and
evaluates deformation and stability through simulation probes. Robot
pick-and-place subsequently tests grasping, transport, and release, providing
interaction feedback for further refinement. A shared physics-grounded harness
governs generation, evaluation, and revision throughout, producing assets with
validation evidence and simulated interaction data.}
\label{fig:pipeline}
\end{figure*}

\section{Method}
\label{sec:method}
\topic{Overview.}
Given a text description or a single image, we seek to construct a deformable
asset for simulated robot interaction. We target volumetric deformable solid
objects; fluids and granular materials are outside our scope. We use the input
to infer geometry and appearance and establish physical models, then refine
them using simulation feedback.
The resulting asset comprises a particle-based physical representation, Gaussian appearance,
and physical configuration, accompanied by validation records and interaction data.
Particle masses, rest volumes, material parameters, and contact conditions
govern the asset's simulated response. Fig.~\ref{fig:pipeline} illustrates the hierarchical asset
generation process, guided by physical probes and robot interaction feedback.

\subsection{Hierarchical Asset Generation}
Generating a deformable asset from text or a single image requires translating
visual and semantic information into a physically grounded representation. We
organize this process into four hierarchical layers, denoted L0--L3, with each
layer building on the outputs of the preceding layers. L0 reconstructs 3D
geometry, L1 establishes a physical model, L2 configures material
properties, and L3 uses robot pick-and-place feedback to adapt manipulation
actions and guide further material revision. This layered structure allows each
part of the asset to be constructed and checked before it supports subsequent
decisions. Later layers use simulation feedback to revise material or action
choices while preserving the geometry and physical model established
earlier.

\topic{3D geometry reconstruction and alignment (L0).}
Asset construction begins with a geometric and visual description that can
support both simulation and rendering. Text input is first converted to a
reference image; image input enters directly. L0 reconstructs the segmented
object's mesh and Gaussian appearance, estimates camera and object geometries, and constructs a convex-decomposed collision proxy for the physical-model
checks in L1. The mesh, collision proxy, sampled particles,
and Gaussians are aligned in a common canonical coordinate frame. Each Gaussian
is associated with neighboring particles in the rest configuration, allowing
simulated particle motion and local deformation to update its position and
orientation. L0 provides aligned geometry for physical modeling and
simulation.

\topic{Physical modeling and initialization (L1).}
Turning this reconstruction into a physical object requires assigning material
volume and mass to the particles and defining the simulation conditions. L1
initializes particle positions and volumes $V_i$ in metric units and assigns
masses $m_i=\rho V_i$ using a uniform density prior $\rho$ proposed by the
L1 Designer from the input description. It also sets gravity,
ground contact parameters, and zero initial velocities.
Placement, drop, and slide tests use this proxy to
temporarily treat the object as rigid and validate its physical model
before deformable simulation.

The resulting physical model is shared by L2 and L3. Mass and volume
enter the material dynamics, while contact conditions determine support and
frictional interaction. Simulation probes and robot interaction use this
common physical model. A benefit of holding these quantities fixed during downstream
material configuration is that it prevents a candidate from compensating for
a different mass, scale, or support condition.
Changes to the physical model require renewed downstream evaluation.

\topic{Deformable material configuration (L2).}
The physical model next requires a constitutive response consistent with the
requested material intent. We assume a homogeneous, isotropic neo-Hookean
material. L2 proposes candidates for Young's modulus, Poisson's ratio, and
damping, then simulates their responses using the material point method
(MPM)~\cite{mpm}. MPM tracks material states on particles and updates their
motion through transfers to and from a background grid.

Simulation probes provide evidence for selecting and refining these candidates.
Deformable drop tests assess impact deformation, rebound, and energy dissipation.
Depending on object geometry, compression or lifting probes assess deformation
under controlled loading and the subsequent response after release.
A separate, longer rollout checks stability.
Numerically and physically valid candidates are ranked against the
requested behavior. The accepted material, together with the L1 physical model,
supplies the deformable asset for robot interaction.

\topic{Robot manipulation (L3).}
Robot manipulation introduces requirements beyond the responses examined by basic physical probes.
L3 inherits the physical model and accepted material, plans a pick-and-place
interaction, and uses its execution to examine grasping, transport, and release.
Interaction feedback guides action and material revisions. Revised actions are
re-evaluated in L3, while material changes require revalidation in L2.
The stage also produces simulated interaction
data alongside the evaluated asset.

\subsection{Robot Interaction and Evidence Generation}
\label{sec:contact}
Grasping and transport require the fingers to establish contact, support the
deforming object through friction, and release it at the destination. To examine
these requirements during generation, the robot stage plans approach, closure,
lift, transport, release, and retreat with the accepted asset. Robot execution
provides the physics harness with contact and deformation observations.

\topic{Contact modeling and force estimation.}
The gripper acts as a kinematic mesh boundary. At an MPM grid node near the
gripper surface, let $u_n$ and $\mathbf{u}_t$ denote the normal and tangential
components of the grid velocity relative to the gripper surface.
To prevent the object from penetrating the gripper, we correct the grid velocity
when its relative normal component points into the gripper ($u_n<0$):
\begin{equation}
 \mathbf{v}_g^{+} = \mathbf{v}_b+\alpha\mathbf{u}_t,
 \label{eq:contact_velocity}
\end{equation}
where $\alpha$ is the fraction of tangential relative velocity retained after
friction,
\begin{equation}
 \alpha = \max\!\left(0,1-\frac{\mu(-u_n)}{\max(\|\mathbf{u}_t\|,%
\varepsilon_v)}\right).
 \label{eq:contact_friction}
\end{equation}
Here $\mathbf{v}_g^{+}$ is the corrected grid velocity,
$\mathbf{v}_b$ is the local gripper surface velocity, $\mu$ is the
gripper--object friction coefficient, and $\varepsilon_v>0$ prevents division
by zero. The grid update blocks motion into the gripper and applies friction
while allowing the object to separate from it. After grid-to-particle transfer,
we correct residual particle penetration and inward relative velocity.
Contact and friction support
the object without attaching it to the gripper.

We estimate the mean reaction force on finger $i$ by summing its contact
impulses over the reporting interval $\Delta t_k$ and dividing by the interval duration:
\begin{equation}
 \overline{\mathbf{f}}_{i}^{\,(k)}
 = -\frac{1}{\Delta t_k}\sum\nolimits_{c\in C_i^{(k)}}\mathbf{J}_c.
 \label{eq:reaction}
\end{equation}
Here $C_i^{(k)}$ contains the finger's contact updates during interval $k$, and
$\mathbf{J}_c$ is the impulse imparted to the object by contact update $c$.
The minus sign gives the opposite reaction on the finger.

\topic{Interaction data.}
We record robot commands, particle states, contact observations, and task
outcomes from each simulated robot interaction. Recorded particle trajectories
drive Gaussian rendering for visual inspection without rerunning the simulation.
We validate recording and replay independently of task success and retain
failed attempts as labeled diagnostic evidence.

\subsection{Shared Physics-Grounded Harness}
\label{sec:harness}
As shown in Fig.~\ref{fig:pipeline}, the shared physics-grounded harness
coordinates proposal, evaluation, and revision across layers L0--L3. Geometric checks,
physical probes, and robot interactions provide evidence for accepting a
candidate or guiding the next revision.

\topic{From proposals to evidence.}
The Planner selects a permitted action or revision route, and the Designer
proposes a structured candidate. The Prober evaluates the candidate and returns
evidence; the Critic interprets it to recommend acceptance or revision.
The Planner, Designer, and Critic are LLM agents, while the Prober performs
geometric checks or simulation and the Orchestrator applies rule checks against
the harness contract. The Orchestrator accepts the candidate or returns feedback
to the Planner for another iteration within the revision budget.
The Prober collects state, contact, and deformation observations from simulation
rollouts, while geometry evaluation uses aligned views and geometric checks.
The recorded evidence connects each decision to the intervention that produced it.

\topic{Harness contract.}
The harness contract defines each stage's permitted actions, required
observations, hard gates, editable fields, and revision budget.
Revisions must stay within the editable fields and parameter bounds and pass
all hard gates before being ranked by agreement with the requested behavior
under fixed test conditions. Each attempt records
its observations, proposal source, and revision history. Only accepted candidates
replace the current version; the search ends when the revision budget is exhausted.

\topic{Manipulation-guided refinement.}
\label{sec:revision}
Basic physical probes do not fully capture an asset's behavior during grasping,
transport, and release. Robot interaction therefore provides task-specific
evidence for further refinement. As shown in the lower-right panel of
Fig.~\ref{fig:pipeline}, the harness uses this feedback to diagnose action,
material, and numerical issues.

Action revisions adjust grasp selection, closure, and motion timing.
Numerical revisions tune simulation hyperparameters while preserving action
duration. Material revisions address undesired deformation after
action diagnosis and require L2 revalidation followed by robot confirmation
under the same commanded plan, physical model, and force budget.
The harness enforces a force budget for each finger during robot interaction.

% Queue the qualitative comparison before the fixed results tables.
\begin{figure*}[!t]
\centering
\vspace*{2.5mm}
\includegraphics[width=.98\textwidth]{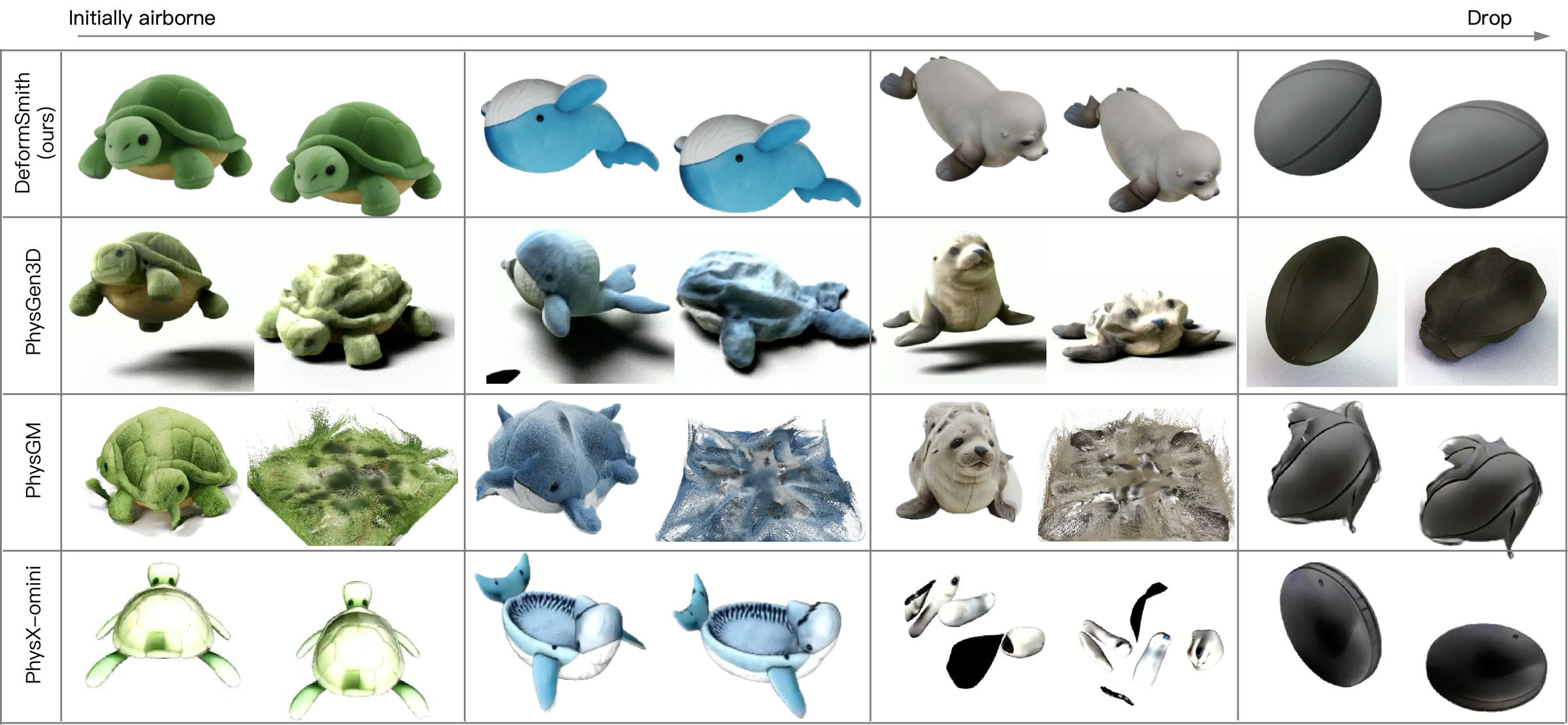}
\caption{\textbf{Qualitative comparison of deformable asset generation and drop simulation.}
Four cases (turtle, whale, seal, and rugby ball) are shown, with each pair
depicting the initial airborne state and a subsequent drop-simulation snapshot.
Compared with PhysGen3D~\cite{physgen}, PhysGM~\cite{physgm}, and
PhysX-Omni~\cite{physxomni}, DeformSmith more consistently
preserves recognizable appearance and coherent object structure.}
\label{fig:assets}
\end{figure*}

\section{Implementation Details}
\label{sec:implementation}
\topic{Reconstruction and simulation.}
We use Qwen-Image-2512~\cite{qwen} to generate reference images from text,
SAM~3~\cite{sam3} for object segmentation, and SAM~3D~\cite{sam3d} for mesh
and Gaussian reconstruction. MoGe-2~\cite{moge} supplies camera and metric
geometry estimates for alignment, and CoACD~\cite{coacd} decomposes the mesh
into convex collision components. Deformable probes and robot interaction use
the Warp MPM backend integrated from DeformMaster~\cite{deformmaster}.
We use particle-driven Gaussian deformation inspired by SC-GS~\cite{scgs}.
We use SAPIEN~\cite{sapien} for robot simulation with a provided URDF
of the RealMan RM65-6F robot. The robot scene is reconstructed with
3D Gaussian Splatting~\cite{gs,real2simeval}.
We use GraspNet~\cite{graspnet} to generate grasp poses from point clouds.

\topic{Harness.}
The current implementation uses GPT-5.6 Sol~\cite{gpt56sol} by default for
the Planner, Designer, and Critic with role-specific prompts. Each role receives
relevant task context, current configurations, stage constraints, and probe
evidence. Responses follow predefined JSON schemas. The harness checks each
response against the role's permitted operations and parameter ranges before
applying configuration changes or running the requested probes.

\section{Results}
\label{sec:experiments}

\subsection{Experimental Setup}
The experiments evaluate deformable asset construction quality and the
benefits of hierarchical generation (C1), the shared physics-grounded
harness (C2), and manipulation-guided refinement (C3). Specifically, we test
whether the proposed hierarchy and feedback mechanisms improve asset quality
and enable generated assets to satisfy both material requirements and
manipulation objectives.

\topic{Data preparation.}
We evaluate 39 cases: 30 text-driven cases and 9 image-based cases.
The text-driven set focuses on volumetric deformable objects suitable for
robot grasping. The image-based set contains 9 target objects
from the public project assets of PhysGen3D~\cite{physgen}.

\topic{Baselines.}
We compare with several state-of-the-art image-to-3D/4D methods, including
PhysGen3D~\cite{physgen}, PhysGM~\cite{physgm}, and
PhysX-Omni~\cite{physxomni}.
Internal ablations examine hierarchical construction, the shared physics-grounded
harness, and manipulation-guided refinement.

\topic{Metrics.}
To support a more rigorous evaluation, we use GPT-6 Astra~\cite{gpt6astra},
a more capable model than the GPT-5.6 Sol used in our harness, to rate physical
realism, photorealism, and semantic consistency on a 0--1 scale, with the input image and task
description as references, following PhysGen3D~\cite{physgen}.
These automated ratings are complemented by blinded pairwise human comparisons
of physical plausibility, visual quality, and semantic consistency, following
SceneSmith's preference protocol~\cite{scenesmith} and adapting PhysGen3D's
perceptual criteria~\cite{physgen}.
Metrics for the ablation studies are detailed in the corresponding sections.

\begin{figure*}[t]
\centering
\vspace*{2.5mm}
\includegraphics[width=.98\textwidth]{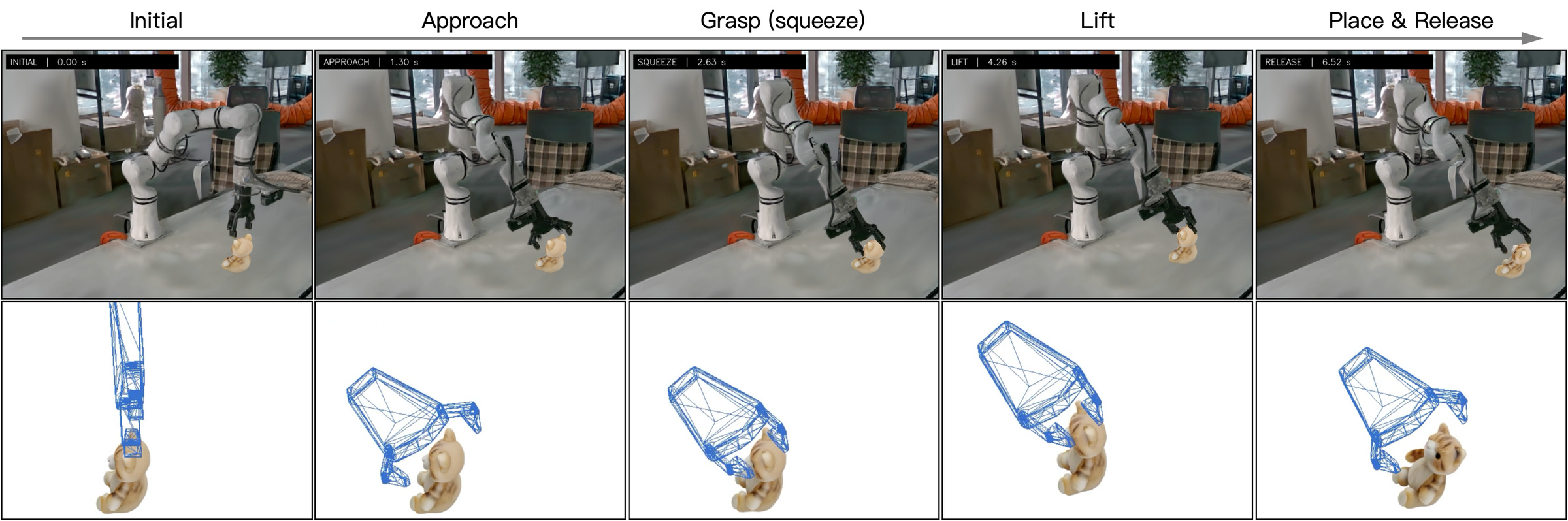}
\caption{\textbf{Simulated robot manipulation with a DeformSmith-generated asset.}
From left to right, a plush object undergoes approach, grasping, lifting, and
release. Scene views (top) are paired with
close-ups (bottom) that highlight object deformation under gripper contact;
the blue wireframe denotes the gripper collision geometry. Dynamic Gaussian
splatting renders the robot scene and deformable object throughout the simulated
interaction, enabling the generation of manipulation data for deformable objects.}
\label{fig:replay}
\end{figure*}

% Keep Tables I and II together below the qualitative figure.
\begin{table}[H]
\centering
\caption{\textbf{Deformable asset quality.} Mean GPT-6 Astra ratings
(0--1; higher is better).}
\label{tab:gpt_quality}
\small
\setlength{\tabcolsep}{3pt}
\begin{tabular}{@{}l|ccc@{}}
\toprule
\raisebox{0.5\normalbaselineskip}{Method} & \shortstack{Physical\\realism $\uparrow$} &
\shortstack{Photo-\\realism $\uparrow$} &
\shortstack{Semantic\\consistency $\uparrow$} \\
\midrule
PhysGen3D~\cite{physgen} & 0.46 & 0.34 & 0.80 \\
PhysGM~\cite{physgm} & 0.39 & 0.27 & 0.68 \\
PhysX-Omni~\cite{physxomni} & 0.41 & 0.32 & 0.69 \\
\textbf{DeformSmith (ours)} & \textbf{0.70} & \textbf{0.58} & \textbf{0.82} \\
\bottomrule
\end{tabular}

\par\vspace{1em}
\caption{\textbf{Deformable asset quality.} DeformSmith win rates in human preference comparisons (\%, ties excluded).}
\label{tab:assets}
\small
\setlength{\tabcolsep}{3pt}
\begin{tabular}{@{}l|ccc@{}}
\toprule
\raisebox{0.5\normalbaselineskip}{Comparison} & \shortstack{Physical\\plausibility $\uparrow$} &
\shortstack{Visual\\quality $\uparrow$} &
\shortstack{Semantic\\consistency $\uparrow$} \\
\midrule
Ours vs. PhysGen3D & 68\% & 88\% & 63\% \\
Ours vs. PhysGM & 73\% & 95\% & 71\% \\
Ours vs. PhysX-Omni & 75\% & 96\% & 76\% \\
\bottomrule
\end{tabular}
\end{table}

\subsection{Complete Asset Construction}
DeformSmith achieves the highest ratings across all three criteria in
Table~\ref{tab:gpt_quality}. Physical realism reaches 0.70 and photorealism
reaches 0.58, each exceeding the strongest baseline, PhysGen3D, by 0.24.
The gain in semantic consistency is smaller (0.82 versus 0.80).
Human comparisons in Table~\ref{tab:assets} show a consistent preference for
DeformSmith: win rates range from 68\% to 75\% for physical plausibility,
88\% to 96\% for visual quality, and 63\% to 76\% for semantic consistency.
Together, these results show that the clearest improvements concern visual
quality and physical behavior, while maintaining agreement with the requested
interaction.

The comparisons use matched input images and a shared interaction description:
a five-second gravity drop onto a horizontal table from a clearance of one
quarter of object height, recorded at 30 fps from two views. Videos share
physical duration and playback speed. All 40 participants independently rated
every pair with anonymized outputs and randomized sides. Physical plausibility
concerns deformation, contact, and recovery; visual quality concerns shape,
appearance, and artifacts. Semantic consistency measures agreement with the
interaction description. Win rates exclude ties and weight inputs equally.
Pairwise evaluation requires viewable outputs from both methods, including
failed dynamics. Inputs to GPT-6 Astra also omit method labels.

Fig.~\ref{fig:assets} illustrates the visual differences underlying these
ratings. Across the turtle, whale, seal, and rugby ball, DeformSmith retains
recognizable shape, surface appearance, and coherent structure in the drop
snapshots. PhysGen3D produces recognizable initial assets but exhibits pronounced
flattening or collapse after the drop. PhysGM shows severe spreading and
surface disruption in the plush cases, while PhysX-Omni exhibits incomplete
geometry, including a fragmented seal. These examples help explain why
recognizable initial appearance alone is insufficient for high asset quality:
the reconstructed object must also remain coherent as it deforms under contact.

This combination of visual fidelity and plausible response is consistent with
DeformSmith's hierarchical construction and shared physics-grounded harness.
L0 aligns reconstructed geometry, Gaussian appearance, and simulation particles,
so physical motion can update the rendered object coherently. L1 establishes mass, volume, and contact conditions before L2 selects material
parameters through deformation probes and stability checks. The harness uses
these observations to reject invalid candidates and guide revision under fixed
physical conditions. The results support the effectiveness of this integrated
pipeline for constructing visually coherent, deformable assets. They assess
perceived quality under the tested interaction, without establishing
material-parameter accuracy.

Fig.~\ref{fig:replay} illustrates a generated asset undergoing squeezing,
lifting, and release during simulated robot manipulation. The asset deforms
under gripper contact while retaining a coherent shape and appearance across
the sequence. These interactions are governed by the contact formulation in
Sec.~\ref{sec:contact} (Eqs.~\ref{eq:contact_velocity}--\ref{eq:contact_friction}),
which couples the deformable asset to the moving gripper through nonpenetration
and friction while allowing separation during release. This example supports DeformSmith's
integration of asset construction with robot interaction, allowing the generated
asset to be exercised under the contact and loading conditions of its intended use.

\subsection{Ablation Studies}
\label{sec:materialeval}
\label{sec:ablations}
\topic{Hierarchical structure (C1).}
Table~\ref{tab:material}-A shows that hierarchical construction improves asset
delivery from 83\% to 93\% and independent physics-test success from 80\% to
87\%, with comparable simulation effort.
The comparison uses the same inputs, initial geometry, models, feedback, and
computation limits. \emph{Hierarchical} constructs geometry, physical models,
and materials progressively, whereas \emph{Flat} can revise them jointly.
Both variants undergo the same final tests, with missing assets counted as
failures.

The gains in both delivery and physics pass rate suggest that staged
construction helps produce assets that are usable and physically valid.
Establishing geometry and physical conditions before material refinement gives
later decisions a consistent basis and constrains the scope of each revision.
The improvement at comparable simulation cost supports hierarchical
organization as a useful component of asset construction.

\topic{Physics-grounded harness (C2).}
In Table~\ref{tab:material}-B, the harness raises material-target satisfaction
from 40\% to 73\% and reduces hard failures from 17\% to 7\%.
Target satisfaction requires basic physical validity and agreement with the
specified material behavior, such as stiffness and recovery, in independent
simulation tests. Hard failure denotes a failure of basic physical validity,
including numerical instability, severely implausible deformation, or missing
valid outputs. The rates are not complementary: a physically valid asset may
still fail to satisfy the target material behavior.
Both variants start from the same physical model and initial material proposal.
\emph{One-shot} retains that proposal, while \emph{Harness} refines it using
simulation feedback, averaging 14 construction calls per asset. Both are
assessed on independent deformation and recovery tests held out from candidate
selection. The zero calls for One-shot refer to construction, excluding the
shared final evaluation.

The simultaneous improvement in target satisfaction and failure rate indicates
that refinement improves agreement with the requested material behavior while
reducing invalid outcomes. Simulation probes expose how a proposed material
actually deforms and recovers, giving the harness evidence to guide subsequent
revisions. These results support the shared physics-grounded harness as an
effective refinement procedure, with the gains reflecting both structured
feedback and the additional simulation effort.

\begin{table}[t]
\centering
\vspace*{2.5mm}
\caption{\textbf{Ablation studies of DeformSmith.}
(A) Hierarchical versus flat asset construction.
(B) Material refinement with the shared physics-grounded harness versus
one-shot prediction. (C) Manipulation-guided refinement versus no feedback.
Joint success requires both task completion and satisfaction of the material
requirements. Simulation calls report the average construction cost per asset.}
\label{tab:material}
\small
\setlength{\tabcolsep}{3pt}
\begin{tabular*}{0.95\columnwidth}{@{\extracolsep{\fill}}l|ccc@{}}
\toprule
\multicolumn{4}{l}{\textbf{A. Hierarchical structure}}\\
\raisebox{0.5\normalbaselineskip}{Variant} & \shortstack{Asset\\delivery $\uparrow$} &
\shortstack{Physics\\pass $\uparrow$} & \shortstack{Sim.\\calls} \\
\midrule
Flat & 83\% & 80\% & 13.7  \\
\textbf{Hierarchical} & \textbf{93\%} & \textbf{87\%} & 13 \\
\midrule
\multicolumn{4}{l}{\textbf{B. Shared physics-grounded harness}}\\
\raisebox{0.5\normalbaselineskip}{Variant} & \shortstack{Target\\satisfaction $\uparrow$} &
\shortstack{Hard\\failure $\downarrow$} & \shortstack{Sim.\\calls} \\
\midrule
One-shot & 40\% & 17\% & 0 \\
\textbf{Harness} & \textbf{73\%} & \textbf{7\%} & 14 \\
\midrule
\multicolumn{4}{l}{\textbf{C. Manipulation-guided refinement}}\\
\raisebox{0.5\normalbaselineskip}{Variant} & \shortstack{Task\\success $\uparrow$} &
\shortstack{Material\\pass rate $\uparrow$} & \shortstack{Joint\\success $\uparrow$} \\
\midrule
No feedback & 40\% & 67\% & 27\% \\
\textbf{Full} & \textbf{67\%} & \textbf{83\%} & \textbf{57\%} \\
\bottomrule
\end{tabular*}
\end{table}

\topic{Manipulation-guided refinement (C3).}
Table~\ref{tab:material}-C shows that the full refinement procedure increases
pick-and-place success from 40\% to 67\%, material pass rate from 67\% to
83\%, and joint success from 27\% to 57\%. Joint success requires a successful
robot trial and an asset that passes the independent material tests.
Task and joint rates are measured over robot trials, while material pass rate
is measured over assets. The comparison uses six assets with five held-out
conditions each, starting from the same L2 assets. \emph{No feedback} retains
the initial assets and action rules; \emph{Full} uses robot feedback to revise
actions and materials, with material changes revalidated before testing.

The increase in joint success shows that improved manipulation performance is
accompanied by better satisfaction of the material requirements. Robot
interaction tests whether the asset can sustain grasping, transport, and
release, exposing requirements beyond those assessed by basic material probes.
Using this feedback to refine both the asset and its interaction supports
manipulation-guided refinement as the final stage of construction.
Asset and action choices are frozen before held-out evaluation, so the results
assess the complete refinement procedure, including its additional search
effort.

\subsection{Real-World Application}
\label{sec:realrobot}
\begin{figure}[t]
\centering
\vspace*{2.5mm}
\includegraphics[width=.85\linewidth]{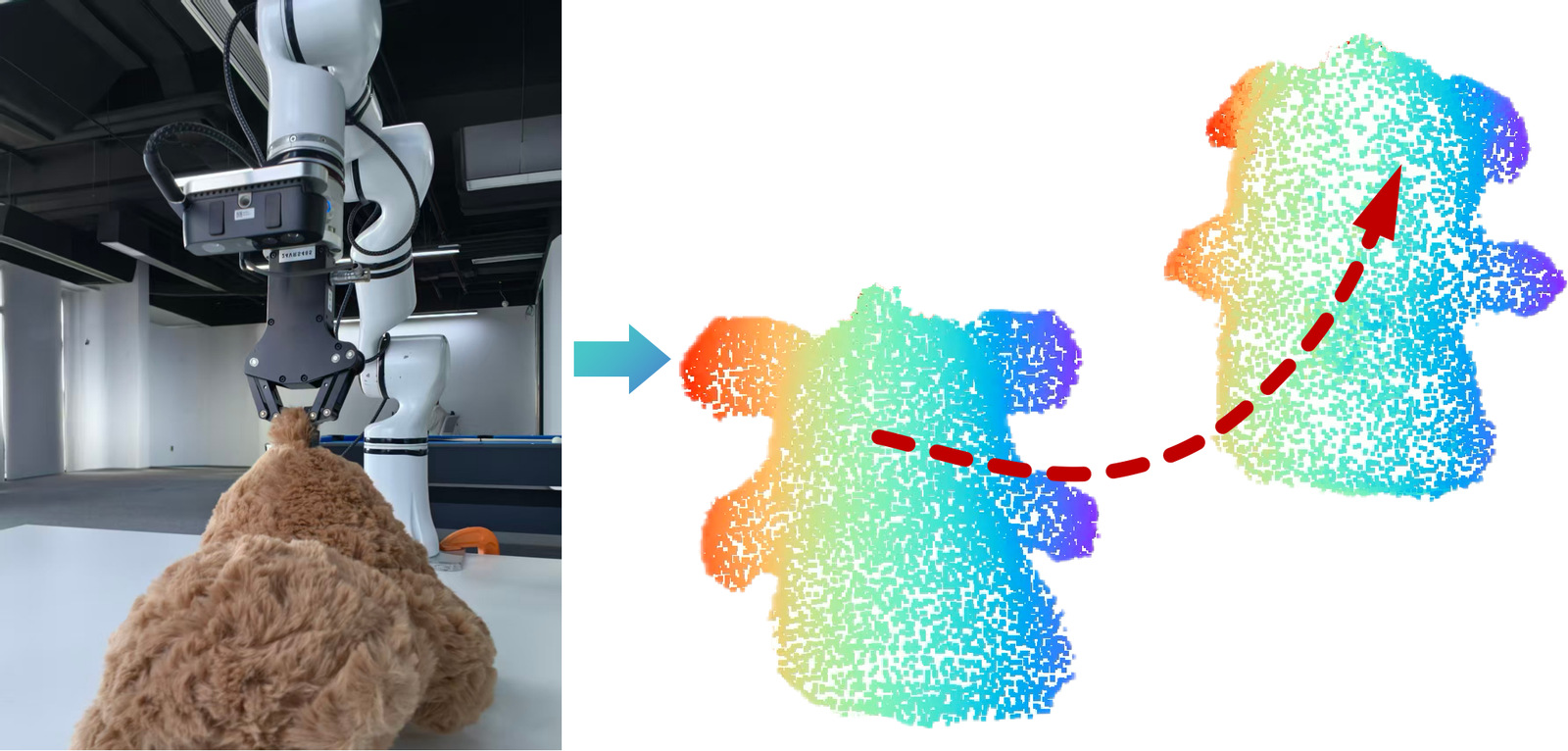}
\caption{\textbf{Potential use of DeformSmith for real-world material parameter estimation.}
DeformSmith-generated geometry and physical parameters could serve as an
initial estimate for subsequent material parameter estimation.
Real robot manipulation (left) provides RGB-D video from which 3D point tracks
are extracted (right), supplying observations that could guide refinement of
this initial estimate.}
\label{fig:realrobot}
\end{figure}

DeformSmith could provide an initialization for modeling real deformable
objects from interaction observations. In principle, its generated geometry
and physical configuration could be combined with observation-driven modeling
approaches such as DeformMaster and EMPM~\cite{deformmaster,empm}.
Starting from this initialization, 3D point tracks extracted from RGB-D video
could constrain material parameter refinement, for example by adjusting
Young's modulus to reduce discrepancies between simulated deformation and
observed motion. Fig.~\ref{fig:realrobot} illustrates the real robot
interaction and motion observations relevant to this potential application.

\section{Conclusion}
We presented DeformSmith, a physics-harness-guided generative framework for
constructing deformable assets from text or a single image. Experiments show improved physical plausibility,
visual quality, and semantic consistency over state-of-the-art baselines.
Ablation studies further demonstrate the effectiveness of hierarchical
construction, the shared physics-grounded harness, and manipulation-guided
refinement.

The current formulation assumes homogeneous volumetric materials and
approximate robot contact, limiting its coverage of complex deformable objects.
Physical parameters inferred from text or images also require real-world
validation, motivating future work on video-based identification and broader
material and contact models.

\bibliographystyle{bibtex/IEEEtran}
\bibliography{references}
\end{document}